%% file: neurips_2022.tex
\documentclass{article}
\pdfoutput=1

\usepackage[nonatbib,final]{neurips_2022}

\usepackage[utf8]{inputenc} 
\usepackage[T1]{fontenc}    
\usepackage[backref]{hyperref}       
\usepackage{url}            
\usepackage{booktabs}       
\usepackage{amsfonts}       
\usepackage{nicefrac}       
\usepackage{microtype}      
\usepackage{graphicx}
\usepackage{amsmath}
\usepackage{amsthm}
\usepackage{booktabs}
\usepackage{algorithm}
\usepackage{algorithmic}
\usepackage{tabularx}
\usepackage{multirow}

\theoremstyle{definition}
\newtheorem{definition}{Definition}[section]

\usepackage{xcolor}

\title{
Towards True Lossless Sparse Communication in Multi-Agent Systems
}

\author{%
  Seth Karten\thanks{Correspondence: \texttt{skarten@cs.cmu.edu}}\\ 
  Carnegie Mellon University\\
  \And 
  Mycal Tucker\\ 
  Massachusetts Institute of Technology\\
  \And 
  Siva Kailas\\ 
  Carnegie Mellon University\\
  \And 
  Katia Sycara\\ 
  Carnegie Mellon University\\
}

\begin{document}

\maketitle

\begin{abstract}
Communication enables agents to cooperate to achieve their goals. Learning \textit{when} to communicate, i.e., \textit{sparse} (in time) communication, and \textit{whom} to message is particularly important when bandwidth is limited. Recent work in learning sparse individualized communication, however, suffers from high variance during training, where decreasing communication comes at the cost of decreased reward, particularly in cooperative tasks. We use the information bottleneck to reframe sparsity as a representation learning problem, which we show naturally enables lossless sparse communication at lower budgets than prior art. In this paper, we propose a method for \textbf{true lossless sparsity} in communication via \textit{Information Maximizing Gated Sparse Multi-Agent Communication} (IMGS-MAC). Our model uses two individualized regularization objectives, an information maximization autoencoder and sparse communication loss, to create informative and sparse communication. We evaluate the learned communication `language' through direct causal analysis of messages in non-sparse runs to determine the range of lossless sparse budgets, which allow \textbf{zero-shot sparsity}, and the range of sparse budgets that will inquire a reward loss, which is minimized by our learned gating function with \textbf{few-shot sparsity}. To demonstrate the efficacy of our results, we experiment in cooperative multi-agent tasks where communication is essential for success. We evaluate our model with both continuous and discrete messages. We focus our analysis on a variety of ablations to show the effect of message representations, including their properties, and lossless performance of our model.
\end{abstract}

\input{sections/01_introduction}
\input{sections/02_related_work}

\input{sections/03_problem_setup}

\input{sections/04_proposed_methodology}
\input{sections/05_experiments}
\input{sections/06_conclusion}

\bibliographystyle{abbrv}
\bibliography{bib}

\clearpage
\input{sections/07_appendix_setup}

\end{document}

%% file: sections/01_introduction.tex
\section{Introduction}
In multi-agent teams, communication is necessary to successfully complete tasks when agents have partial observability of the environment. Multi-agent reinforcement learning (MARL) has recently seen success in scenarios that require communication~\cite{foerster2016learning,commnet,freed2020simultaneous,lowe2017multi, lazaridou2016multi}. 
Sparse multi-agent communication (wherein agents communicate during only some time-steps) has been shown to be an effective solution to internet packet routing~\cite{mao2020learning}, multi-robot navigation~\cite{benSparseDiscrete}, complex multiplayer online games such as StarCraft~\cite{commnet,ic3net,jiang2018learning,foerster2016learning}, and human-agent teaming~\cite{karten2021inter}. In particular, these successes have been achieved using neural network architectures in conjunction with a reinforcement learning framework.
Simultaneously, research in individualized multi-agent communication~\cite{commnet,ic3net,tarmac,graphMA} has solved sparse cooperative-competitive multi-agent problems where adversaries are listening, and sparsity is built into their competitive objective. But such research is unable to provide sparse individualized communication in fully-cooperative settings, where there is no built-in incentive. This is particularly unreasonable in real-world settings where multiple robots may need to adhere to bandwidth/budget restrictions. A budget (or bandwidth) $b$ defines the maximum percentage of the time an agent may communicate.

\begin{figure}[!t]
    \centering
    \includegraphics[width=0.75\textwidth]{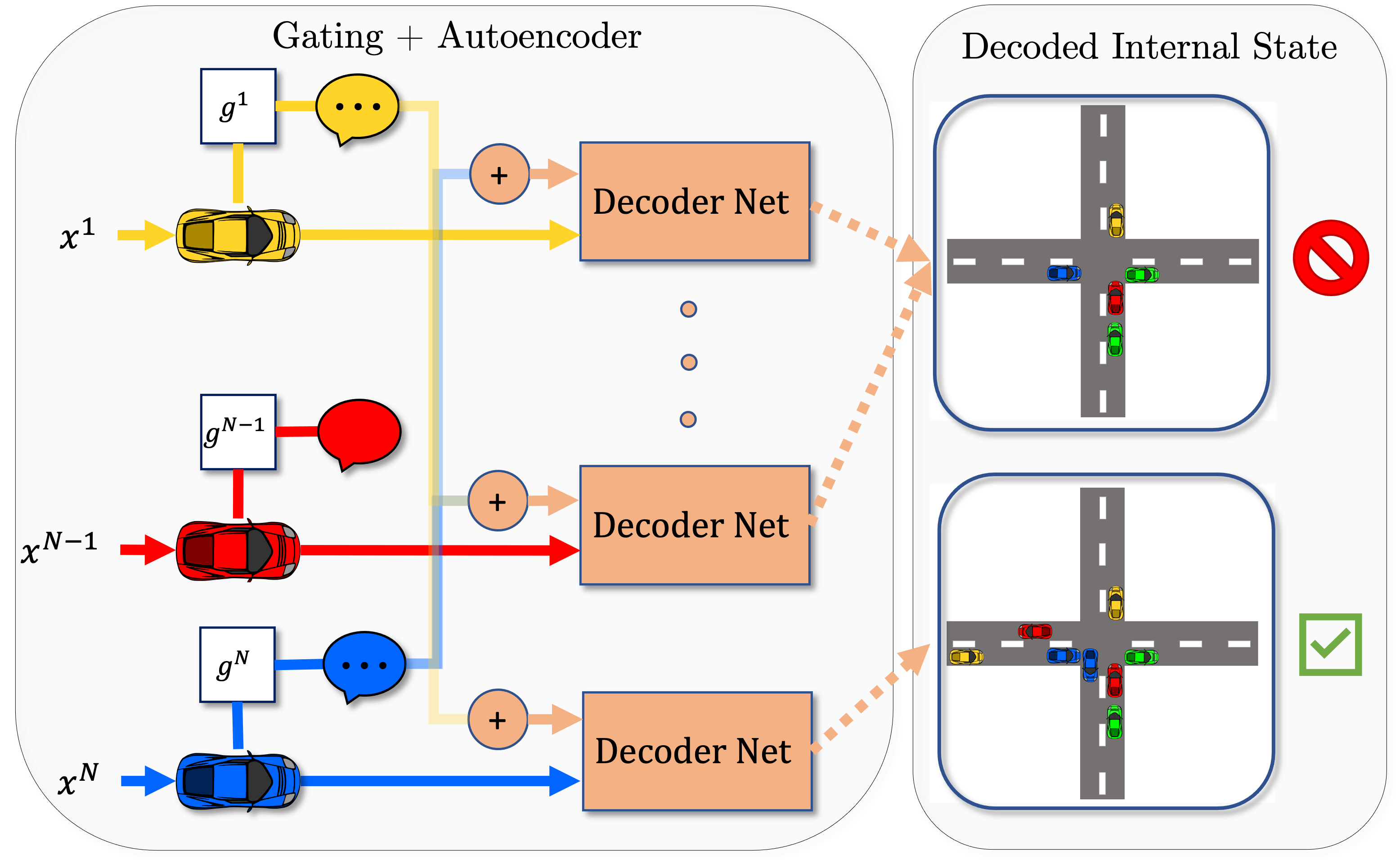}
    \caption{Overview of our multi-agent architecture with gated sparse, informative communication. At every timestep, each agent receives an occluded observation $x$. Each agent creates a communication message, which is passed to the learned gating function $g$ as well as the $\texttt{Decoder}$. The gating function determines whether to communicate the message to the other agents. The $\texttt{Decoder}$ receives all messages and attempts to reconstruct the full state of the environment.}
    \label{fig:FIG1}
\end{figure}


Emergent communication enables agents to learn a set of communications vectors apt for solving a particular task; however, learning emergent communication simultaneously with an action policy is highly unstable. Agents often converge to undesirable policies in which communication is ignored, unless special training terms are used~\cite{EcclesBiases, MA_autoencoder}. Enforcing sparse communication, i.e., limiting the number of messages over time or communicating within a bandwidth/budget, only worsens this problem due to the additional constraint. 
Using the information bottleneck framework \cite{tishby2015deep} may adequately address sparsity constraints~\cite{wang2020learning}, but due to their objective, exhibit a trade-off between the total bandwidth and task performance. 
In these scenarios, the agents fail to send necessary messages and transmit unnecessary messages, which we dub \textit{null communications}. In fact, many papers on sparsity suggest lossless sparsity, but in actuality, have a non-trivial decrease in reward. 

In this work, we propose a novel framework, \textit{Information Maximizing Gated Sparse Multi-Agent Communication} (IMGS-MAC), which aims to learn a communication-action policy and then enforce a sparse communication budget (learning \textit{when} and \textit{whom} to send messages) with lossless performance. 
Our key insight in IMGS-MAC is reframing the sparse multi-agent communication problem as a representation learning problem. 
The use of an information maximizing autoencoder prevents shortcut solutions in order to structure the latent communication space to allow for high reward with little communication. 
After learning a non-sparse communication policy, we analyze the direct causal effect of choosing to send each token to any other agent to determine null messages.
Then, IMGS-MAC 
uses a table of these null messages to prevent them from being emitted, enabling sparsity with lossless performance without additional reinforcement learning, which we call zero-shot sparsity.
To further promote sparsity for over-constrained budgets, we finetune our model using an individualized communication regularization term for a learned gating/targeting function $g$, which we call few-shot sparsity. 

%% file: sections/02_related_work.tex
\section{Related Work}\label{related}

\subsection{Emergent Communication Vectors}
Prior art in emergent communication establishes how agents may learn to communicate to accomplish their goals with continuous communication vectors~\cite{jiang2018learning,commnet,ic3net}. Motivated by human communication in which people speak only when necessary and using only a discrete set of words, we wish for agents to learn sparse (in number of listeners over time) and discrete communication. While previous work has been successful in learning discrete communication vectors~\cite{lazaridou2020emergent,li2021learning,foerster2016learning,agrawal2021learning,benSparseDiscrete}, the learned communication conventions often exhibit undesirable properties.
Learning discrete prototypes has been shown to promote robustness in noisy communication channels, as well as human interpretability and zero shot generalization \cite{discreteComm}. Similar to word embeddings in natural language processing, they capture the relationship between vectors. However, many of these methodologies only try to learn token meanings through rewards. In our work, we show that grounding messages in reproducing the concatenated state of all agents with an autoencoder creates desirable representations regardless of continuous or discrete settings.

\subsection{Sparsity: Gating Total Messages}
In this work, we attempt to reduce communication in MARL problems through gating total messages.
Gating methods learn a function which dictates whether an agent will communicate to each other agent at any given timestep.
Some methods try to learn a gating probability to decide whether to broadcast a budget, but these are unable to follow a communication budget~\cite{ic3net,ETCNet}.
In reward-based sparse communication~\cite{karten2021inter,vijay2021minimizing}, by penalizing communication reward during training, gating/targeting methods have reduced communication. However, this method is not able to adequately choose a budget (what maximum percentage of the time to communicate). Overall, gating methods are high variance and often unstable~\cite{agrawal2021learning}.
Rather than building the objective into the reward, I2C~\cite{I2C} tries to measure the causal effect of an individualized message through a learned Q-value. However, I2C only tries to address sparse targeting in the lossless sparsity case and fails to account for the effect of message representation. In our work, we measure the actual effect of each token and mask the emergent vocabulary accordingly.

%% file: sections/03_problem_setup.tex
\section{Preliminaries}\label{problem}

We formulate our setup as a centralized training, decentralized execution (CTDE)~\cite{foerster2016learning}, partially observable Markov Decision Process with individualized communication (POMDP-Comm). 
Formally, our problem is defined by the tuple, $(\mathcal{S},\mathcal{A},\mathcal{M},\mathcal{T},\mathcal{R},\mathcal{O},\Omega,\gamma)$. We define $\mathcal{S}$ as the set of states, $\mathcal{A}_i \, , \, i\in[1,N]$ as the set of actions, which includes task specific actions, and $\mathcal{M}_i$ as the set of communications for $N$ agents.  $\mathcal{T}$ is the transition between states due to the multi-agent joint action space $\mathcal{T}: \mathcal{S} \times \mathcal{A}_1,...,\mathcal{A}_N \to \mathcal{S}$. $\Omega$ defines the set of observations in our partially observable setting. The partial observability requires communication to complete the tasks successfully. $\mathcal{O}_i: \mathcal{M}_1,...,\mathcal{M}_N \times \mathcal{S} \to \Omega$ maps the communications and state to a distribution of observations for each agent. $\mathcal{R}$ defines the reward function and $\gamma$ defines the discount factor.

\subsection{The Sparsity Objective}
The multi-agent emergent communication problem is phrased as a combination of a Lewis game~\cite{lewis1969convention} and the information bottleneck~\cite{tishby2015deep}.
We seek to develop a message representation $M$, which contains sufficient referential and ordinal information to successfully complete a task.
Notably, the information bottleneck defines a trade-off between referential ($X$) mutual information, $I(X; M)$, which is observable to an agent, and ordinal ($Y$) mutual information, $I(M; Y)$, which requires coordination between agents.

The communication graph $G_t = (V,E)$ is a set of agents (vertices) and active communication edges between them, where connectivity changes at each timestep. Messages flow through the edges from agents to agents, $E: v_i \to v_j$. We aim to learn a masking function $g$ to dynamically modify the graph to prevent messages from flowing along the graph.
The total number of bits communicated, $s(M)$ can be defined in terms of vertices (gating), $v \in V$, $s(M) = \sum_{m\in \mathcal{M}} v_m$ or in terms of edges (targeting), $e \in E$, $s(M) = \sum_{m\in \mathcal{M}} e_m$, over an episode. One can see that gating is a special form of targeting in which a vertex is disjoint from the graph. We will use gating and targeting interchangeably, but in terms of sparsity, limit the total number of message edges during an episode.

In MARL, the objective of sparse communication is to minimize the total number of bits communicated while maximizing team task performance,
\begin{equation}\label{eq:max_min_sparse}
    \begin{aligned}
        &\max\limits_{\pi: \mathcal{S} \to \mathcal{A} \times \mathcal{M}} \mathbb{E} \left[  \sum_{t \in \mathcal{T}} \sum_{i \in N} \gamma \mathcal{R}(s_t, a_t)  \right]\\
        &\text{s.t. } (a_t, m_t) \sim \pi, s_t \sim \mathcal{T}(s_{t-1}) \\
        &\text{subject to}\\
        &\min \mathbb{E}_{M \sim \pi}\left[ s(M) \right]
    \end{aligned}
\end{equation}
That is, to achieve this objective, first one maximizes task performance; then one reduces total communication while keeping task performance fixed.

\begin{definition}[Lossless Sparse Communication]
A communication policy $\pi_m$ is lossless and sparse iff it satisfies the objective in equation \ref{eq:max_min_sparse}. A lossless sparse communication policy defines the minimum sparse budget (fraction of total messages) $b^*$.
\end{definition}

Most sparse communication work rephrases the $\min \max$ problem to a single objective by introducing a Lagrangian,
\begin{equation}\label{eq:sparse_obj}
    \begin{aligned}
        &\max\limits_{\pi: \mathcal{S} \to \mathcal{A} \times \mathcal{M}} \mathbb{E} \left[  \sum_{t \in \mathcal{T}} \sum_{i \in N} \gamma \mathcal{R}(s_t, a_t)  - \lambda s(m_t) \right]\\
        &\text{s.t. } (a_t, m_t) \sim \pi, s_t \sim \mathcal{T}(s_{t-1}), m_{AVG} < b
    \end{aligned}
\end{equation}
However, depending on the Lagrange multiplier, the objective in equation~\ref{eq:max_min_sparse} is not always the same as equation~\ref{eq:sparse_obj}. 
Due to the dual-objective, equation \ref{eq:sparse_obj} also introduces the possibility of suboptimal sparse communication even when lossless sparse communication is possible. It also explains the high variance of lossless sparsity in prior art~\cite{agrawal2021learning}.

\begin{definition}[Sub-Optimal Sparse Communication]\label{def:sub}
A communication policy $\pi_M$ is suboptimal and sparse iff there exists a trade-off between task performance and messaging constraints as defined in equation \ref{eq:sparse_obj}.
\end{definition}

Thus, in our methodology, we cannot directly optimize equation~\ref{eq:sparse_obj}. 
Recall that in emergent communication, messages are generated based on their observations. This implies that, in terms of the information bottleneck, messages represent a combination of referential, $I(X;M)$, and ordinal, $I(M;Y)$, information. That is, observations help guide ordinal (task-specific) information. Suppose, we have a Lagrangian objective (see section~\ref{method_sparse}), which allows for our messages to have independent referential information. Then, given a communication policy which adequately solves the task, one can determine the ordinal utility of each token. By removing unnecessary tokens, we can satisfy the objective in equation~\ref{eq:max_min_sparse}.
Thus, in our methodology, we emphasize learning emergent communication with properties that enable sparse communication with lower optimal budgets $b^*$ (lossless sparsity).

%% file: sections/04_proposed_methodology.tex

\section{Proposed Methodology}\label{methodology}
\begin{algorithm}[!t]
\caption{IMGS-MAC}\label{alg:cap}
\begin{algorithmic}[1]
\STATE $\theta \gets \text{randomly initialized network parameters}$
\STATE $\texttt{useDiscreteMessaging} \gets \{true | false\}$
 \WHILE{\textit{not converged}}
 \FOR{$i\gets 1 \text{ to } N$ \COMMENT{simultaneously} } 
    \STATE $x^i \sim \mathcal{S}$
    \STATE $h^i \gets \texttt{GRU}(x^i)$
    \IF {$\texttt{useDiscreteMessaging}$} 
        \STATE $m^i \gets \texttt{DiscreteProtoNet}(h^i)$
    \ELSE
        \STATE $m^i \gets h^i$
    \ENDIF
    \STATE $\texttt{SendMessages}(m^i \odot g(h^i))$
    \STATE $\Bar{m}^i \gets \texttt{AggregateMessages}()$
    \STATE $\Tilde{h}^i \gets \texttt{GRU}(\{h^i , \Bar{m}^i\})$
    \STATE $a^i, v^i, \Tilde{x}^i \gets \pi(\Tilde{h}^i), V(\Tilde{h}^i), \texttt{DecoderNet}(\Tilde{h}^i)$
    \STATE $L \gets \pi\texttt{Loss}(a^i, v^i) + \mathcal{L}_1(x, \Tilde{x}^i) + \mathcal{L}_2(m^i_{AVG})$
\ENDFOR
\ENDWHILE
\end{algorithmic}

\end{algorithm}

In this section, we introduce the IMGS-MAC architecture as well as two types of individualized regularization. The first is an autoencoder, which is used to stabilize the dual training of the communication-action policy. The latter is an individualized communication penalty to enforce each agent individually follows a fixed communication budget/bandwidth. Note that it is important to provide individualized regularization, as otherwise the gradient signal will not be adequately recognized. Our model builds on related art~\cite{ic3net,agrawal2021learning}, but our technique can be easily applied to any individualized MARL communication module. 
Below, we introduce our information maximization autoencoder and individualized communication regularization. Overall, the combined framework can be observed in Alg.~\ref{alg:cap}.

\subsection{Sparsity through Information Maximization}\label{method_sparse}
The information bottleneck principle~\cite{tishby2015deep} is naturally encoded into any communication module that uses deep learning. By creating a latent representation, any nontrivial solution enforces the network to provide the relevant information within the communication vector. 
Rather than requiring centralized execution to maintain sparsity through the information bottleneck, we provide a form of information regularization that allows for individualized communication.
Additionally, we enforce a structured representation for message tokens, ensuring that tokens represent independent referential and ordinal information from their observations.

We define the autoencoder as follows: The communication module of our network serves as the encoder. Each agent produces their own hidden state $h^i$ and receives communication vectors $m^j$ such that $i \neq j$. For each agent, the model feeds $h^i + m^j$ into the decoder. We then calculate the $l2$ loss $U(s_t,s_{t}^{i,\texttt{decoded}}) $ between the state of all agents $s_t = \{x^1_t,\dots,x^N_t\}$ and the decoded state $s_{t}^{i,\texttt{decoded}}$, which effectively measures the similarity between the latent communication and the concatenated state of all agents.
\begin{equation}\label{eq:l1}
    \mathcal{L}_1(\theta) = \lambda_1 U(s_t,s_{t}^{i,\texttt{decoded}}) 
\end{equation}

To enable sparsity, we first train IMGS-MAC with the autoencoder module and non-sparse communication ($b=1$). Afterwards, we run evaluation episodes while collecting data regarding each message token to detect null messages. 
\begin{definition}[Null Communication Vector]
A null communication vector from agent $i$ provides a lack of information to another agent $j$. That is, in terms of the information bottleneck, $I(m^i; y^j) = 0$.
\end{definition}
To determine the mutual information between a message $m$ and the task specific information $y$, we measure if there is a change in the reward within a small $\epsilon \approx 1e-3$. If there is no significant change, we consider this token a null message.

While simple, in our experiments, we show that by combining this trick with strong latent representations, our model can remove larger amounts of unnecessary communication, or null communication vectors, without impacting the performance. In fact, our lossless sparsity method requires no additional reinforcement learning training, which we define as zero-shot sparsity.

Similar to zero-shot learning, which requires no additional data to satisfy an objective, zero-shot sparsity enables satisfaction of sparse communication constraints from non-sparse training through careful analysis of the emergent communication policy.
Our methodology exhibits zero-shot sparsity since no additional reinforcement learning training is required to enforce sparsity given our non-sparse model with informative communication, which is shown in section~\ref{experiments}.

\subsection{Sparsity through Individualized Regularization}
In the overconstrained bandwidth case, $b < b^*$, which implies that we will not be able to maximize task performance, inducing the suboptimal sparsity case. However, we can use the properties of lossless sparsity to maximize performance such that $m_{AVG} <= b$. 
We combine previous techniques with a second regularization term, a per-agent communication penalty $\mathcal{L}_2$. The penalty depends on the nature of the communication budget. At each discrete time-step $t$, each agent has the opportunity to choose to emit a message. Thus, we define our budget $b$ as a fraction of the total agents multiplied by the time-steps in which we measure communications. We let $m_{AVG}$ define the actual fraction of messaging. Finally, we can define the regularization penalty,
\begin{equation}\label{eq:l2}
    \mathcal{L}_2(\theta) = \lambda_2 \bigg\lVert m_{AVG}^i - (b+(1-b^*)) \bigg\rVert_2^2
\end{equation}
where we penalize messages when $b < m_{AVG} < b^*$.

Similar to few-shot learning where a limited amount of data, we define few-shot sparsity as enabling the satisfaction of sparse communication constraints from non-sparse training through limited additional MARL training. We quantify the amount of data in our experiments, notably Fig.~\ref{fig:tri}.
We finetune our model using the regularization penalty in Eq.~\ref{eq:l2} to observe overconstrained budgets, thus exhibiting few-shot sparsity.

%% file: sections/05_experiments.tex
\section{Experiments}\label{experiments}
In this section, we first describe the benchmark environment. Then, we present ablations showing the efficacy of our sparse model with informative communication. 
As stated in section~\ref{related}, IC3Net and I2C provide close framework compatibility. 
We compare IMGS-MAC with IC3Net with non-sparse ($b=1$) communication to understand the effect of our information maximizing autoencoder in developing independent referential (based in observations $x$) representations, $I(m_j; m_k) = 0$. We evaluate with both continuous and discrete messages to show the necessity of using our methodology to develop structured latent tokens (messages $m$). 
Then we show the few-shot sparsity benefits of finetuning sparse budgets when $b<b^*$ as compared with solving the tri-objective (1: communicate effectively, 2: act effectively, and 3: obey communication sparsity constraints), which is akin to trying to satisfy the objective in Eq.~\ref{eq:sparse_obj} when $b \geq b^*$. 
We analyze our model's communication vectors to find zero-shot sparsity $b=b^*$. We show that our method can provide lower optimal budgets $b^*$ than I2C. Finally, we verify that IMGS-MAC has lossless performance at $b=b^*$ as compared with its non-sparse performance $b=1$, and show the optimized trade-off between suboptimal budgets $b<b^*$ and task performance, e.g., reward.
We detail our experimental setup in Appendix~\ref{sec:appendix_setup}.

\subsection{Information Maximization Analysis}
\begin{figure}[!t]
    \centering
    \includegraphics[width=.3\textwidth]{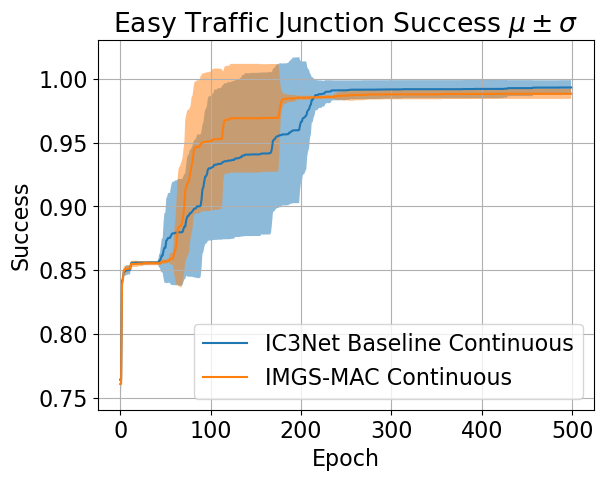}
    \includegraphics[width=.3\textwidth]{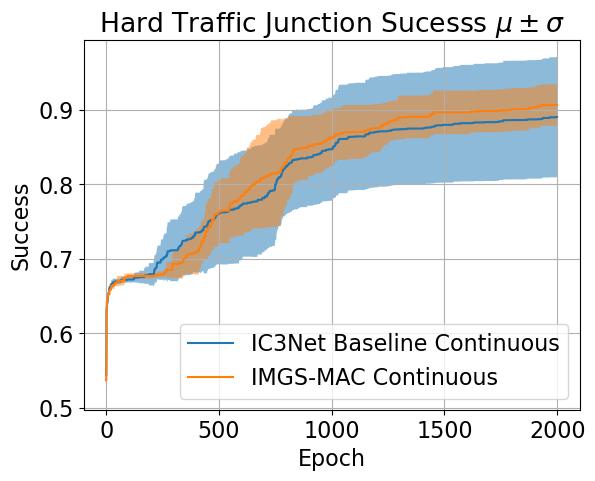}
    \includegraphics[width=.3\textwidth]{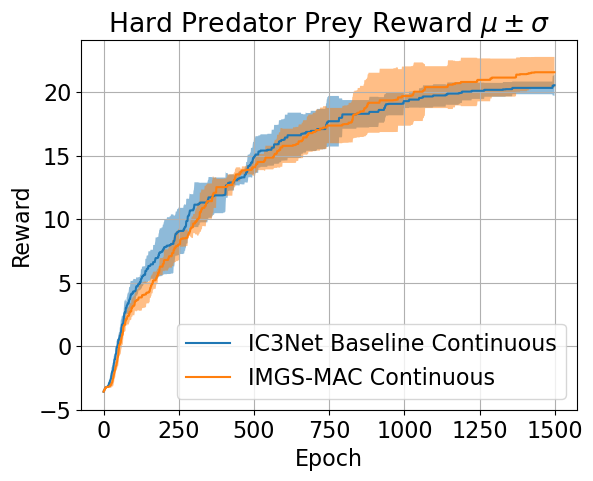}
    \includegraphics[width=.3\textwidth]{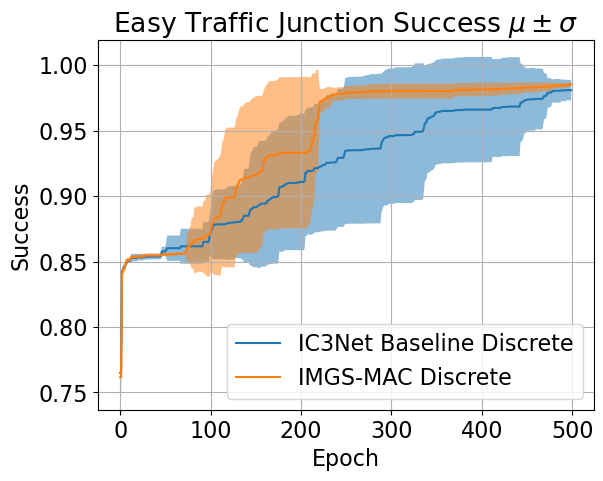}
    \includegraphics[width=.3\textwidth]{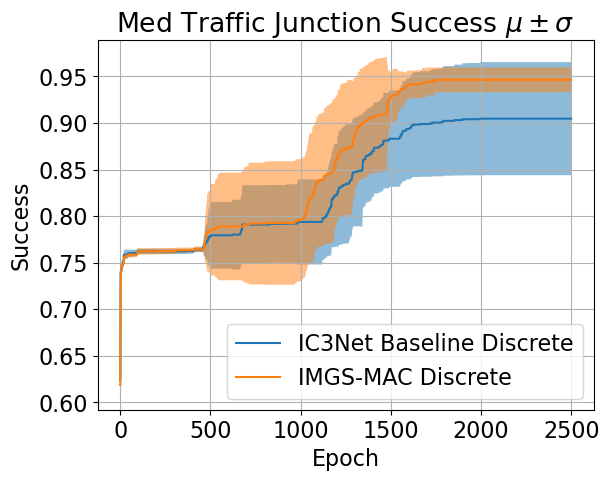}
    \includegraphics[width=.3\textwidth]{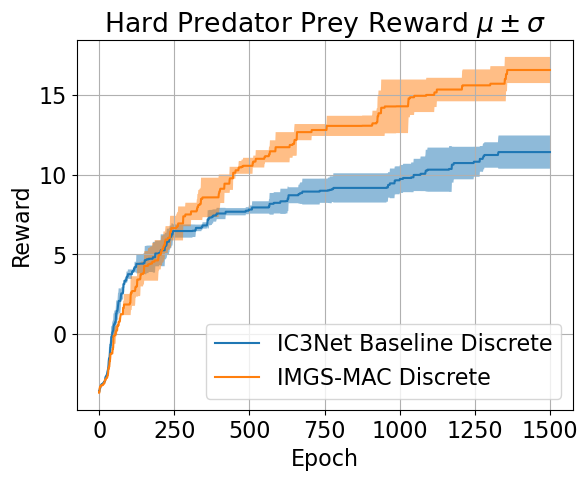}
    \caption{Left and middle figures compare the training IC3Net  (blue) vs. our IMGS-MAC (orange) with non-sparse communication ($b=1$) in Traffic Junction. Our method converges to higher success earlier and with less variance. Right figures compare in Predator-Prey. Our method converges to higher success earlier and with less variance. Top figures use continuous communication vectors while bottom figures use discrete.}
    \label{fig:fixed_baseline}
\end{figure}

To show the benefits of the autoencoder for information maximization, we first show comparison with  IC3Net with a fixed gate, i.e., non-sparse communication ($b=1$). In Figure~\ref{fig:fixed_baseline}, our results show that our method has much lower variance. Note that IC3Net may have a shaded area higher than IMGS-MAC, but it never actually performs that well. Rather, the variance comes from very low performing runs. In the simple, easy setting, our method is able to find solutions of equivalent quality as IC3Net. However, in hard settings, and in all discrete communication vector settings, our method outperforms IC3Net in terms of performance and the number of epochs required to find the solution. Particularly, in the more difficult discrete communication vector scenarios, the autoencoder drastically outperforms IC3Net. Note that the decreased variance results in much more stable solutions.

\begin{table*}[t!]
\centering
    \caption{Average $\mu \pm \sigma$ for quality and performance of null communication vectors. IMGS-MAC (ours) provides significantly more informative communication, as recognized by its low usage of null communications. Lower is better.}
\begin{tabularx}{\textwidth}{X X| X X X} 
 \specialrule{.2em}{.1em}{.1em} 
   \multirow{ 2}{*}{Environment} & \multirow{ 2}{*}{Method} & \% Null Comm. Vectors & \# Observations \newline per Vector & \% Null Comms. \newline Emitted\\
   \hline
   \multirow{ 2}{*}{TJ Easy Cts.} & IC3Net & 0.59 \small $\pm$ 0.107 & 3.81 \small $\pm$ 0.304 & 0.529 \small $\pm$ 0.112\\
      & \textbf{IMGS-MAC} & 0.0550 \small $\pm$ 0.198 & 1.785 \small $\pm$ 0.507 &  0.0565 \small $\pm$ 0.196 \\
   \hline
   \multirow{ 2}{*}{TJ Hard Cts.} & IC3Net & 0.404 \small $\pm$ 0.0753 &  26.892 \small $\pm$ 6.662 &  0.543 \small $\pm$ 0.0999 \\
      & \textbf{IMGS-MAC} & 0.0334 \small $\pm$ 0.107  & 16.928 \small $\pm$ 10.113  &  0.0310 \small $\pm$ 0.167\\
   \hline
    \multirow{ 2}{*}{TJ Easy Discrete} &   IC3Net & 0.589 \small $\pm$ 0.265 & 3.39 \small $\pm$ 1.09 & 0.846 \small $\pm$ 0.263 \\ 


    & \textbf{IMGS-MAC} &0.0194 \small $\pm$ 0.0394 & 1.390 \small $\pm$ 0.220 & 0.0320 \small $\pm$ 0.0719\\
    \hline
    \multirow{ 2}{*}{TJ Med. Discrete} &  IC3Net & 0.724 \small $\pm$ 0.139 & 15.944 \small $\pm$ 8.127 & 0.964 \small $\pm$ 0.0424 \\


    & \textbf{IMGS-MAC} & 0.0857 \small $\pm$ 0.172 & 5.105 \small $\pm$ 3.154 & 0.201 \small $\pm$ 0.322  \\
    \hline
    
    \multirow{ 2}{*}{PP Hard Cts.} &   IC3Net  & 0.784 \small $\pm$ 0.0445 & 73.148 \small $\pm$ 12.099 & 0.497 \small $\pm$ 0.0887 \\ 
    & \textbf{IMGS-MAC} & 0.284 \small $\pm$ 0.160 & 17.523 \small $\pm$ 6.231 & 0.300 \small $\pm$ 0.173\\
    \hline
    \multirow{ 2}{*}{PP Hard Discrete} &  IC3Net  & 0.482 \small $\pm$ 0.145 & 104.803 \small $\pm$ 6.0713 & 0.719 \small $\pm$ 0.312 \\
    & \textbf{IMGS-MAC} & 0.380 \small $\pm$ 0.0909 & 82.809 \small $\pm$ 6.507 & 0.141 \small $\pm$ 0.114  \\
    
\specialrule{.1em}{.05em}{.05em}
\end{tabularx}
    \label{tab:analytical_auto}
\end{table*}

Our hypothesis is that decreasing the training epochs to converge to high task performance implies that we have more informative communication.
Our results show that communication tokens which represent information more independently allow for lower $b^*$. This is found by analyzing the number of states in which the same message is emitted. Overall, this strengthens our hypothesis that a structured latent space naturally allows for lower $b^*$ for lossless sparsity. 
We analytically study the performance of the autoencoder in Table~\ref{tab:analytical_auto}. 

Percent null communication vectors determines the number of null tokens in the emergent `vocabulary', i.e., all possible messages. 
The number of observations per vector reports the independence of a token or mutual information between any two distinct tokens, $I(m_j;m_k)$. We want to minimize $I(m_j;m_k)$ in order to decouple information into independent messages, so that we can later promote stronger sparsity through the analysis of the utility of each token in determining optimal actions.
The percent of null communications emitted reports the percentage of null messages that were communicated to other agents over 500 episodes. We aim to minimize these unnecessary null messages.
We see that the IC3Net method uses more null vectors on average and has high mutual information between tokens. Further, using our IMGS-MAC, we effectively remove null messages and decrease mutual information between tokens, further improving performance. In fact, IMGS-MAC removes almost all null messages.  We will later further see that it does so without any reduction in performance.



\subsection{Sparsity Analysis}

\subsubsection{Few-shot Sparsity}
\begin{figure}[!t]
    \centering
    \includegraphics[width=.24\textwidth]{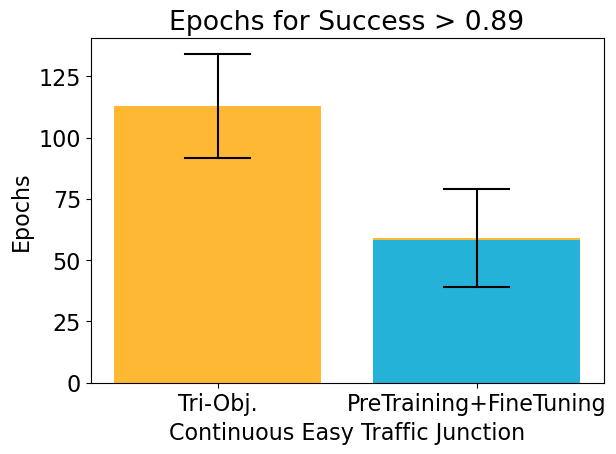}
    \includegraphics[width=.24\textwidth]{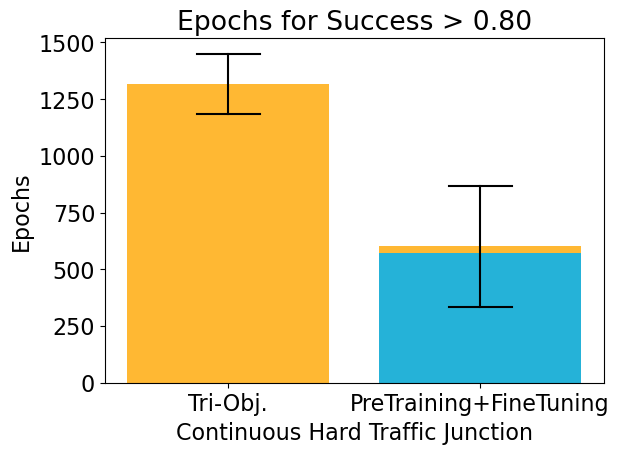}
    \includegraphics[width=.24\textwidth]{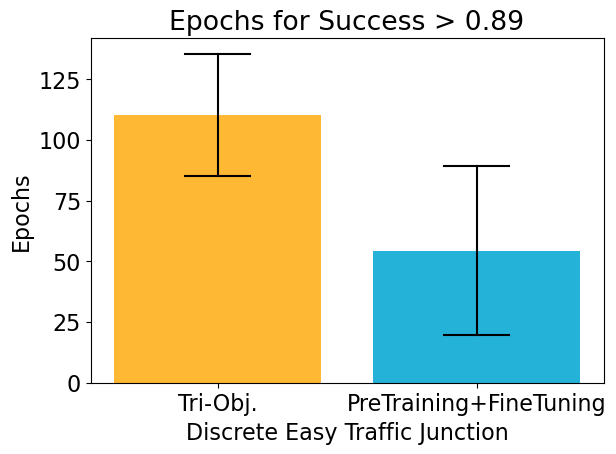}
    \includegraphics[width=.24\textwidth]{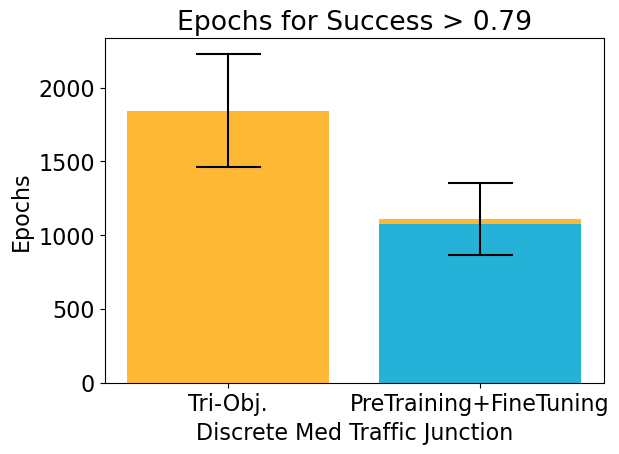}
    \caption{Average success and 95\% confidence interval for Tri-objective (left bar, orange) vs. Pretraining with non-sparse $b=1$ (blue), then Finetuning (orange) with $b=0.7$. The Pretraining+Finetuning paradigm takes half the amount of training as the Tri-objective. }
    \label{fig:tri}
\end{figure}

\begin{figure}[!t]
    \centering
    \includegraphics[width=.24\textwidth]{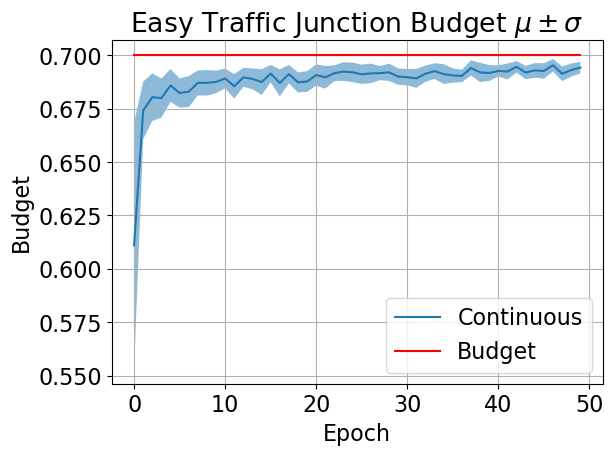}
    \includegraphics[width=.24\textwidth]{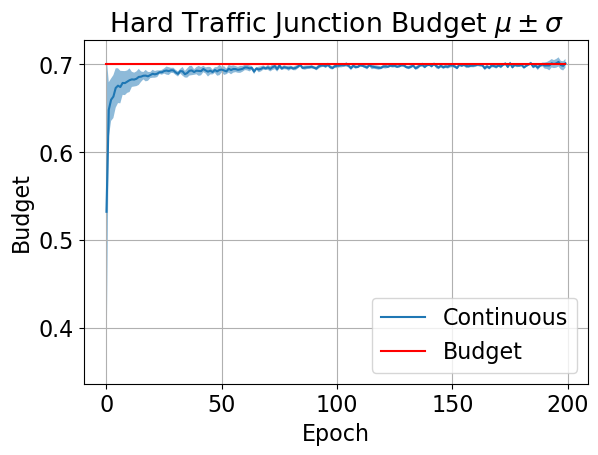}
    \includegraphics[width=.24\textwidth]{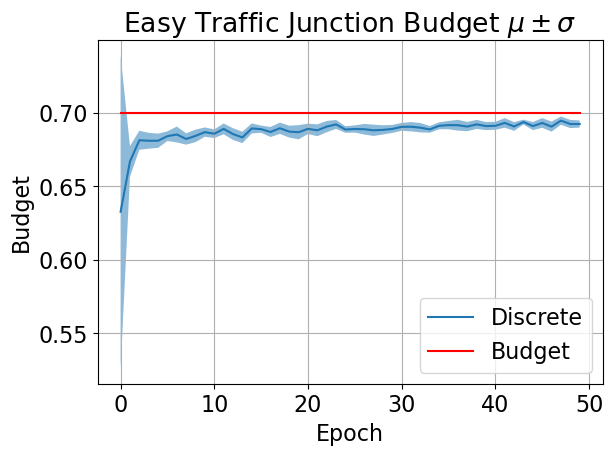}
    \includegraphics[width=.24\textwidth]{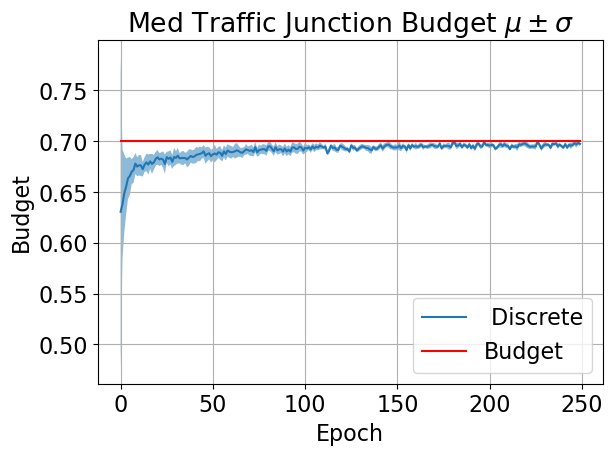}
    \caption{Above, the model follows the budget $b=0.7$ average over each episode. Observe that the model (in blue) only needs to run for a few dozen epochs before adequately following the budget (in red).}
    \label{fig:budgetVar}
\end{figure}
In the case where $b < b^*$ we require a small amount of additional training data to enable sparse communication.
We introduce an autoencoder to include independent referential communication in order to ease the dual communication-action policy learning. When we introduce the sparsity constraint (and the corresponding individualized communication regularization), our model must additionally learn a gating function, which further increases the complexity. 
In order to avoid requiring more data, we introduce a pretraining and finetuning paradigm. First, we pretrain dual communication-action policy with a fixed open gate (non-sparse $b=1$). Then, we apply finetuning to train the gating function (with the rest of the network) at any $b < b^*$.
In Figure~\ref{fig:tri}, we see that the total number of epochs required for task success convergence under a budget is about half as many for the pretraining+finetuning paradigm than for the tri-objective, which aims to solve the objective in Eq.~\ref{eq:sparse_obj} directly.
Note that the variance entirely comes from the dual objective pretraining. The sparsity finetuning requires less than 10\% of the total training epochs. In fact, we can apply finetuning for any budget $b$ rather than having to train the tri-objective from scratch, further decreasing training time.
In Figure~\ref{fig:budgetVar}, we observe that our model only needs a few dozen epochs to converge to a communication budget and is able to safely reduce total communication below the allowed budget.
Overall, our objective exhibits \textit{few-shot sparsity} ($b<b^*$). The performance of few-shot sparsity is analyzed in Figure~\ref{fig:success_vs_budget}.

\begin{table}[t!]
\centering
\caption{Minimum sparse budget $b^*$ with lossless performance, $\mu \pm \sigma$. Observe that our model can reduce 20-60\% without a loss in task performance.}
\begin{tabularx}{.7\textwidth}{|X| X X |} 
 \specialrule{.2em}{.1em}{.1em} 
   
   Environment & \textbf{IMGS-MAC} $b^*$ & I2C-Cts. $b^*$\\ \hline
   TJ Easy Cts. & 0.610 \small $\pm$ 0.191 & - \\
   TJ Hard Cts. &  0.462 \small $\pm$ 0.249 & 0.63 \\
   TJ Easy Discrete & 0.815 \small $\pm$ 0.00469 & - \\ 
   TJ Med Discrete & 0.519 \small $\pm$ 0.140 & 0.66\\ 
   PP Hard Cts. & 0.244 \small $\pm$ 0.0644 & 0.48\\
   PP Hard Discrete & 0.263 \small $\pm$ 0.00757 & 0.48\\
   
\specialrule{.1em}{.05em}{.05em}
\end{tabularx}
\label{table:lossless}
\end{table}

\subsubsection{Zero-shot Sparsity}


We use sparsity through information maximization in section~\ref{method_sparse} to reduce the number and usage of null prototypes.
In Table~\ref{tab:analytical_auto}, one can see that through our analysis, we are able to remove significant usage of null communication vectors, which allow our model to only use informative communication, enabling \textbf{true lossless sparsity}. That is, the task performance, or success in our case, will not decrease at all by decreasing the budget within the true lossless range. Otherwise, enforcing a budget requires the learned gating function $g$ to determine whether an agent should communicate, which may induce a loss in task performance.  Of course, this is dependent on how well the initial communication model is learned, i.e., the range is dependent on the learned model. Each model has its own minimum lossless budget $b^*$, which depends on the emergent communication model. In Table~\ref{table:lossless}, we report the lossless budget $b^*$ for each environment. We are able to reduce communication by 20-75\% with no additional training. Interestingly, we are able to reduce communication more when we have continuous communication vectors instead of discrete communication vectors. This implies that our continuous vectors have more informative communication. Though, it most likely follows from the fact that discrete communication is a harder problem than continuous communication, confirming results from~\cite{tucker2022towards}. 
Additionally, we are able to find lower optimal budgets $b^*$ than I2C, even without specific reinforcement learning training to reduce the communication overhead.

\begin{figure}[!t]
    \centering
    \includegraphics[width=.3\textwidth]{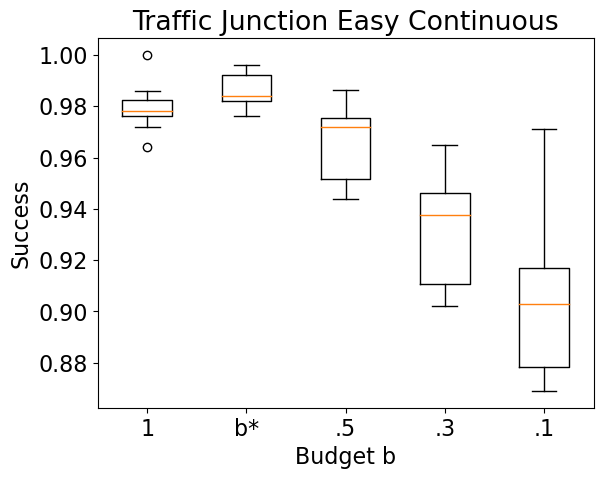}
    \includegraphics[width=.3\textwidth]{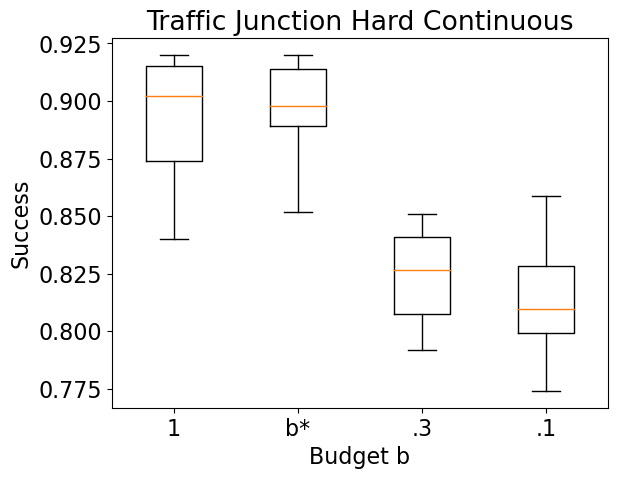}
    \includegraphics[width=.3\textwidth]{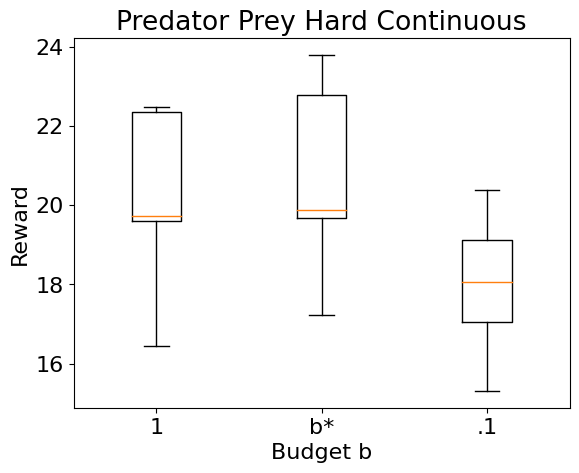}
    \includegraphics[width=.3\textwidth]{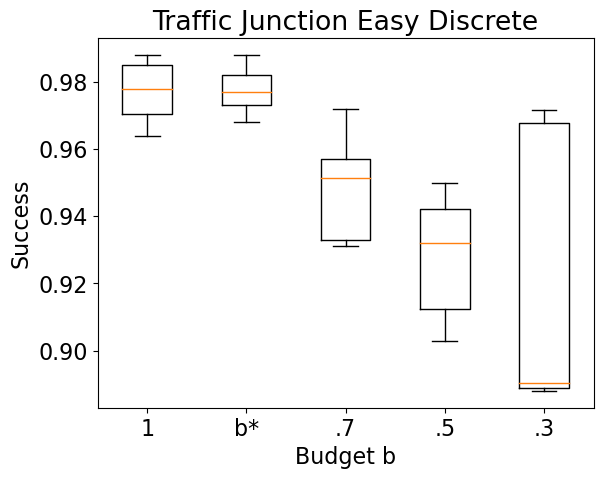}
    \includegraphics[width=.3\textwidth]{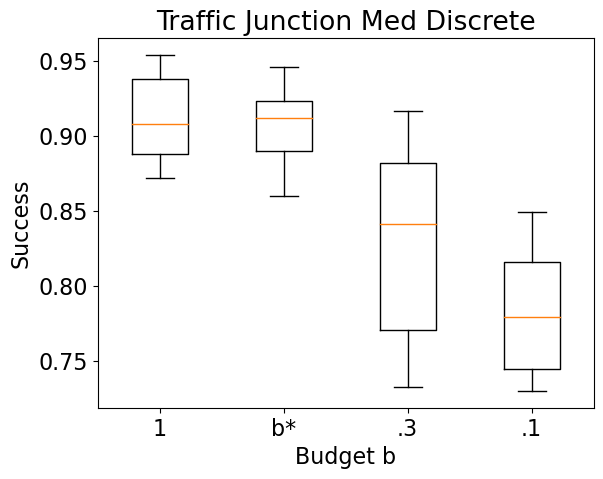}
    \includegraphics[width=.3\textwidth]{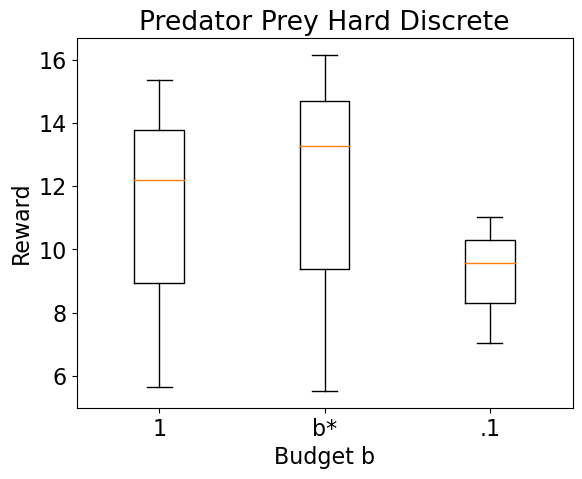}
    
    \caption{Success versus budget for IMGS-MAC at baseline non-sparse $b=1$, lossless $b=b^*$, and suboptimal $b<b^*$. Our model provides lossless performance for $b=b^*$ for $b^*$ in Table~\ref{table:lossless} as compared with the baseline non-sparse $b=1$. Our performance tapers for smaller budgets until it approaches the no communication performance. Top: continuous communication vectors; Bottom: discrete; Left, middle: Traffic Junction; Right: Predator-Prey.}
    \label{fig:success_vs_budget}
\end{figure}

Finally, we analyze the lossless, $b=b^*$, and suboptimal, $b<b^*$, performance for sparse budgets for our model in Figure~\ref{fig:success_vs_budget}, which uses the lossless budget $b^*$ as reported in Table~\ref{table:lossless}. We find that the lossless budget $b^*$ provides true lossless performance. Unsurprisingly, for overconstrained budgets $b<b^*$, there is a small task performance tradeoff for adherence to the budget.


%% file: sections/06_conclusion.tex
\section{Conclusion and Future Work}\label{conclusion}
In this paper, we have proposed a method for multi-agent individualized sparse communication. We reframed sparsity as a representation learning problem through the information bottleneck problem. We have shown that through training a communication-action policy grounded with an autoencoder and analysis during execution of non-sparse messaging, one can exhibit lossless zero-shot sparsity. That is, the sparsity objective may be achieved without any cost of performance with no additional reinforcement learning training. Additionally, we produce individualized regularization to limit performance loss with few-shot sparsity. This allows our model to adhere to messaging constraints in over-constrained bandwidth scenarios.
In a limitation of our work, once the 'vocabulary' is restricted by removing some null messages, other messages are discovered later that could be removed and mutual information between tokens is nonzero. Stronger theoretical bounds on message content independence will further allow sparser communication.
In our future work, we aim to create an overarching framework that combines gating/targeting sparsity and communication compression. This will remove the need for tuning message sizes, but still opt for a decoupled training scenario. That is, first learn an emergent language. Then adhere to sparsity constraints. Additionally, further increases to the unsupervised representation learning will allow for sparser performance. 
\\

%% file: sections/07_appendix_setup.tex
\appendix{}
\section{Experimental Setup}\label{sec:appendix_setup}
\begin{figure}[!t]
    \centering
    \includegraphics[width=.3\textwidth]{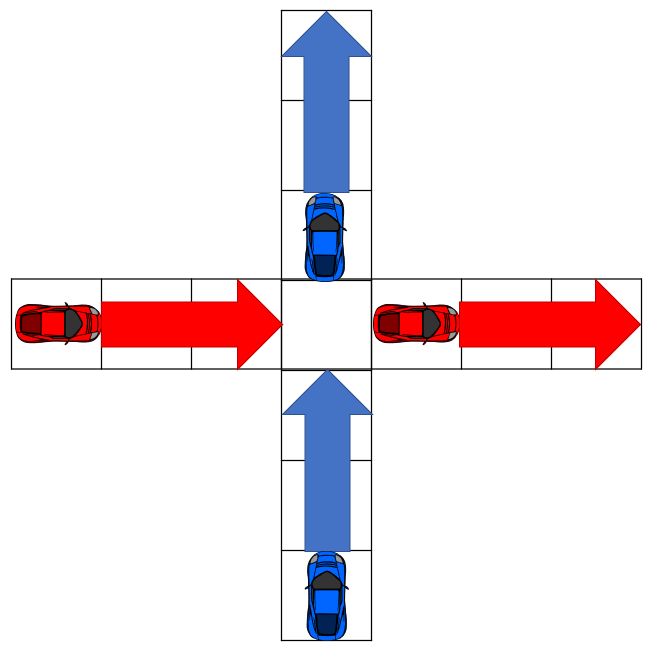}
    \includegraphics[width=.3\textwidth]{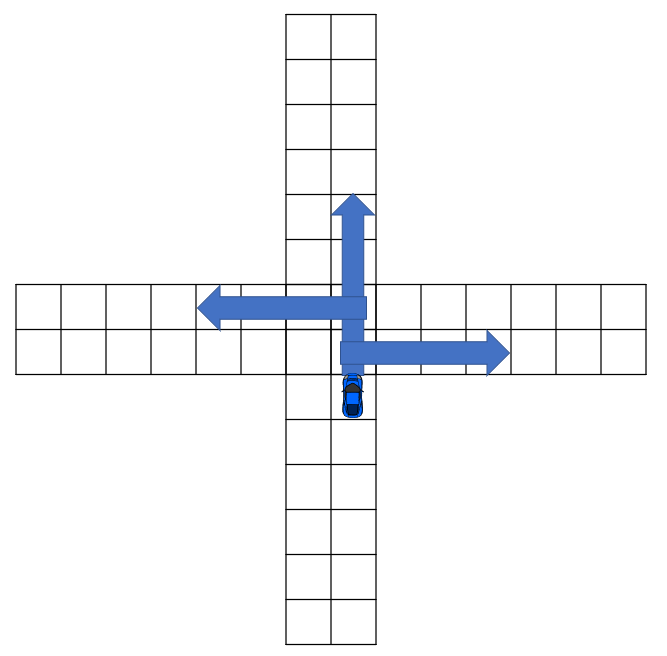}
    \includegraphics[width=.3\textwidth]{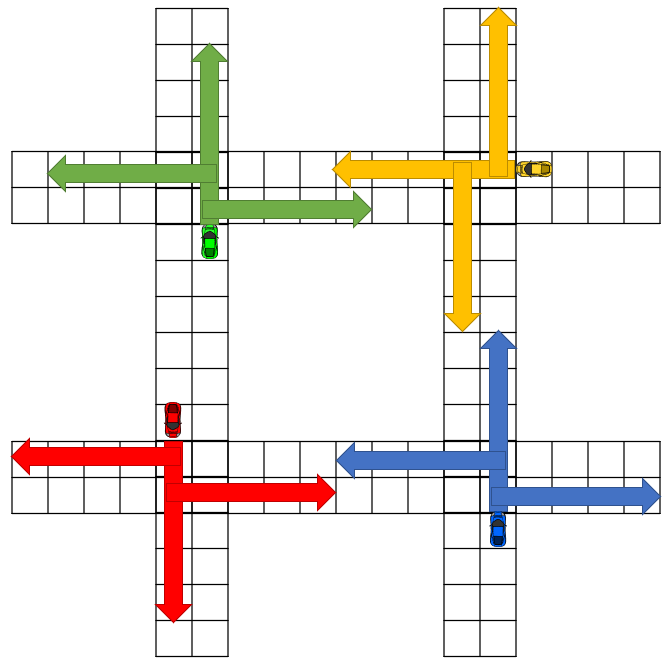}
    \includegraphics[width=.55\textwidth]{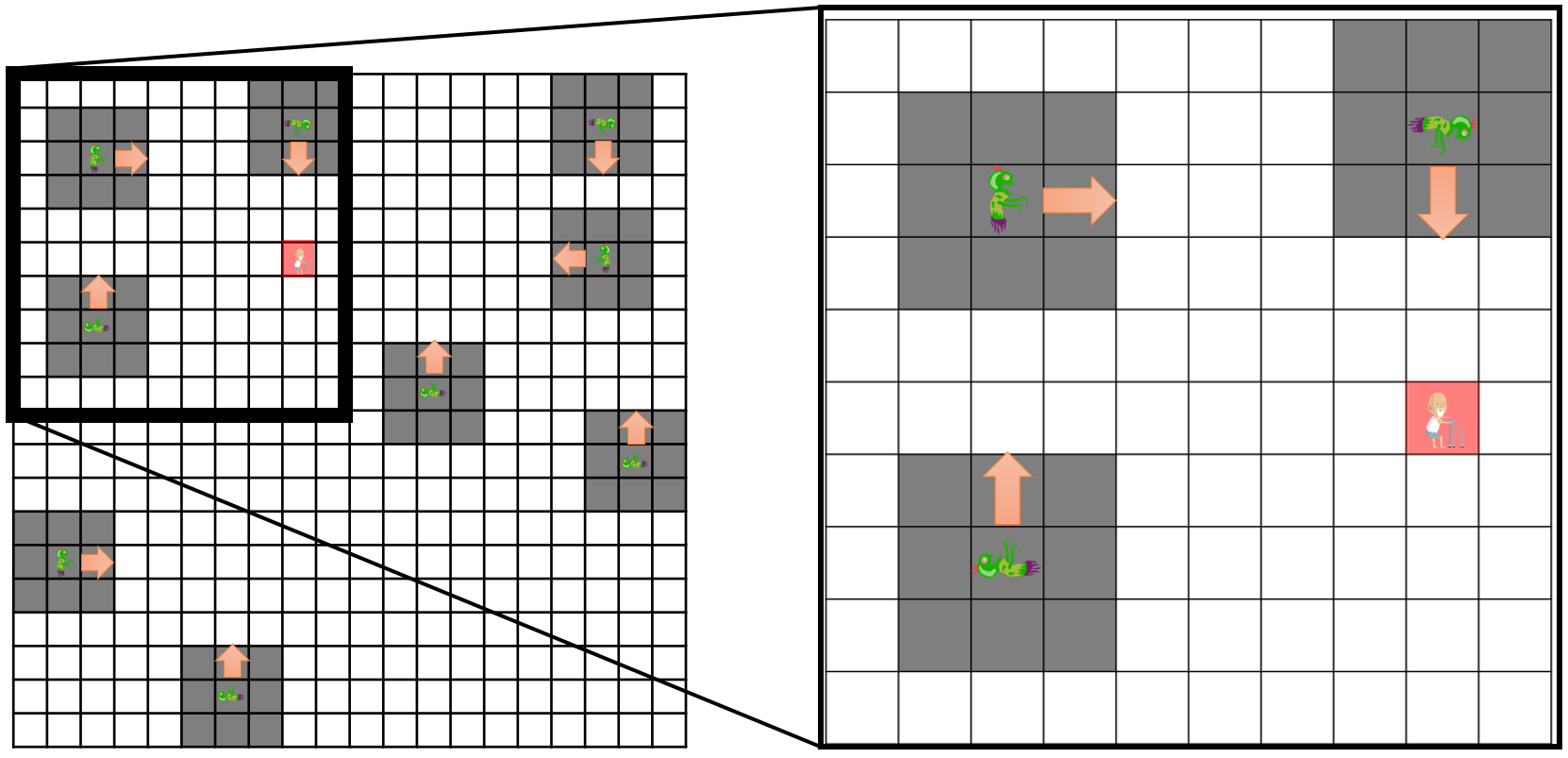}
    \caption{Above are the easy, medium, and hard traffic junction environments. Visibility is limited to the cell the car is located, so agents are effectively blind. The bottom shows a zoomed-in view of the $20 \times 20$ predator-prey environment. The predators are denoted by green aliens, while the prey is denoted by a human (in a red square).}
    \label{fig:env}
\end{figure}
We train and evaluate our model in a blind traffic junction and predator-prey environment settings following prior benchmarks~\cite{ic3net,commnet,I2C}. For each of these variants, we train on 10 random seeds and one epoch uses 5000 samples. We used an RMSProp optimizer with a learning rate of 0.003. See Figure~\ref{fig:env}.

The blind traffic junction scenario involves multiple agents navigating a discretized narrow intersection with no observability regarding the locations of the other agents. Clearly, this necessitates informative communication in order to avoid collisions in the environment. Note that both communication and action occur in a single time-step. 
We study three variants of the blind traffic junction and report results on the easiest and hardest environments which converge for continuous and discrete communication. 

The predator-prey scenario involves multiple agents, where one agent is denoted as the prey and the remaining agents are denoted as predators. The predator agents move and search the environment for the prey agent. The predator agents can only observe its current cell and the adjacent cells (limited visibility to 1 cell around itself). The episode terminates when all predator agents reach the prey agent or when the maximum episode length is hit.

Predator-prey does not necessarily require communication to solve the task. However, in the fully-cooperative predator-prey environment, predators are rewarded for maximizing the number of predators who reach the discovered prey. Thus, there is no built-in incentive for fully-cooperative teams to decrease total communication. In our experiments, we show that our method, IMGS-MAC, is able to decrease messaging to a minimum sparse budget $b^*$ with lossless performance.

Overall, our proposed method is trained (``pretraining'') using the autoencoder in Eq.~\ref{eq:l1}. We then analyze to determine if our model will follow a lossless sparse budget. If not, we finetune our model for a suboptimal sparse budget (Def.~\ref{def:sub}) using the message penalty in Eq.~\ref{eq:l2}.

We use REINFORCE \cite{williams1992simple} to train both the gating function and policy network subject to the previous constraints. In order to calculate the information similarity, we compute loss, using Eq.~\ref{eq:l1}, between each agent's decoded state $s_t^{i,\texttt{decoded}}$ and the concatenation of all agents' states $s_t$.